\documentclass[conference]{IEEEtran}
\IEEEoverridecommandlockouts

\usepackage{cite}
\usepackage{amsmath,amssymb,amsfonts}
\usepackage{algorithmic}
\usepackage{graphicx}
\usepackage{textcomp}
\usepackage{tabularx} 
\usepackage{arabtex}
\usepackage{utf8}
\setcode{utf8}
\usepackage{array}
\usepackage{booktabs}
\usepackage{xcolor}
\usepackage{float}
\def\BibTeX{{\rm B\kern-.05em{\sc i\kern-.025em b}\kern-.08em
    T\kern-.1667em\lower.7ex\hbox{E}\kern-.125emX}}
\begin{document}

\title{An Ensemble Classification Approach in A Multi-Layered Large Language Model
Framework for Disease Prediction}

\author{
\IEEEauthorblockN{{\large Ali Hamdi\IEEEauthorrefmark{1},
Malak Mohamed\IEEEauthorrefmark{1},
Rokaia Emad\IEEEauthorrefmark{1},
Khaled Shaban\IEEEauthorrefmark{2}}}

\IEEEauthorblockA{\IEEEauthorrefmark{1}\large\textit{Dept. of Computer Science, MSA University, Giza, Egypt} \\
\{ahamdi, malak.mohamed17, rokaia.emad\}@msa.edu.eg}

\IEEEauthorblockA{\IEEEauthorrefmark{2}\large\textit{Dept. of Computer Science, Qatar University, Doha, Qatar} \\
khaled.shaban@qu.edu.qa}
}

\maketitle

\begin{abstract}
Social telehealth has made remarkable progress in healthcare by allowing patients to post symptoms and participate in medical consultations remotely. Users frequently post symptoms on social media and online health platforms, creating a huge repository of medical data that can be leveraged for disease classification. Large language models (LLMs) such as LLAMA3 and GPT-3.5, along with transformer-based models like BERT, have demonstrated strong capabilities in processing complex medical text. In this study, we evaluate three Arabic medical text preprocessing methods such as summarization, refinement, and Named Entity Recognition (NER) before applying fine-tuned Arabic transformer models (CAMeLBERT, AraBERT, and AsafayaBERT). To enhance robustness, we adopt a majority voting ensemble that combines predictions from original and preprocessed text representations. This approach achieved the best classification accuracy of 80.56\%, thus showing its effectiveness in leveraging various text representations and model predictions to improve the understanding of medical texts. To the best of our knowledge, this is the first work that integrates LLM-based preprocessing with fine-tuned Arabic transformer models and ensemble learning for disease classification in Arabic social telehealth data.
\end{abstract}


\section{Introduction}
The growth of social telehealth has revolutionized the provision of healthcare, enabling patients to share their symptoms and even consult with doctors remotely . During the COVID-19 pandemic, social telehealth increased greatly in popularity; at that time, access to traditional healthcare services was limited~\cite{hamad2022steducov}. Social media platforms, particularly health forums, are becoming increasingly valuable sources of user-generated medical data that people use to share detailed symptoms and seek the advice of doctors' opinions~\cite{b1,b2,b4}. However, the unstructured and noisy nature of this data poses a great challenge; hence, advanced computational techniques are required for effective analysis. Recent advances in NLP, especially with the advent of large language models, have really transformed unstructured textual data processing~\cite{b1,b2,b4}. Various models, including LLAMA, GPT, and BERT, among others, had promised outstanding performance related to text classification and understanding based on large-scale pre-training over big datasets to obtain advanced on a wide range of domains~\cite{b4,b3,abdellatif2024lmrpa,alali2025claseg,wei2019eda}. With these advances, challenges arise in applying LLMs to domain-specific tasks, including healthcare. The fine-tuning of specialized application LLMs have to consider domain-specific nuances, noisy data handling, and optimizing computational efficiency~\cite{11078357}. This is a resource- and computationally intensive process involving new frameworks so that the effectiveness of LLMs in real-world applications is optimized~\cite{b5}.

\begin{figure*}[h]
    \centering
    \includegraphics[width=0.99\textwidth]{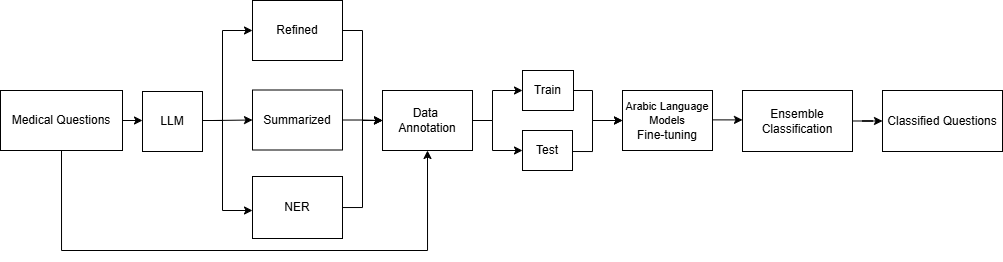}
    \caption{Proposed Ensembling approach in a Multi-Layered Framework for Enhancing Arabic Language Model Fine-Tuning with LLAMA3 Preprocessing}
    \label{fig:image1}
\end{figure*}

We propose a framework that combines LLM-based preprocessing with fine-tuning of Arabic language models for disease classification. It refines text, summarizes posts, and extracts medical entities using NER, enhancing task-specific fine-tuning and overcoming traditional method limitations.

Figure~\ref{fig:image1} illustrates the flow process of our proposed framework. The raw text data generated by users is processed through three LLM-based preprocessing methods: refinement, summarization, and Named Entity Recognition (NER). To preserve contextual richness, the original text posts are combined separately with each preprocessed output, resulting in three paired representations: (original + refined), (original + summarized), and (original + NER). Each paired dataset is then annotated and used to fine-tune Arabic language models. Finally, their predictions are aggregated using a majority voting ensemble for multi-class, multi-label classification tasks, including disease type prediction.

Our contributions in this work are as follows:
\begin{itemize}
    \item \textbf{Novel Integration of LLMs and Arabic Language Models:} We present a multi-layered framework that employs LLMs for preprocessing (refinement, summarization, NER) to enhance the fine-tuning of Arabic language models intended for healthcare applications.
    \item \textbf{Enhanced Preprocessing Pipeline:} We create an improved dataset. LLM-based preprocessing for making challenges related to user-generated content more interpretable and more structured.
    \item \textbf{Improvement of Classification Performance:} We enhanced the accuracies of disease type classification by combining fine-tuned pre-trained Arabic language models like CAMeL-BERT, AraBERT, and Asafaya-BERT on LLM-preprocessed using a majority voting ensemble technique.
\end{itemize}

This work emphasizes the transformative potential of exploiting LLMs in order to enhance fine-tuning for domain-specific language models when handling real-world challenges in healthcare.

\section{Related Work}
Recent breakthroughs in text classification and fine-tuning of large language Models have enhanced NLP applications across domains. In our previous study~\cite{mohamed2025llm}, we fine-tuned CAMeL-BERT, AraBERT, and Asafaya-BERT on the same dataset, but without applying an ensemble strategy. Transformer-based models such as BERT and its variants have achieved remarkable performance on a variety of tasks such as false information detection, sentiment analysis and radical content classification~\cite{b9,b8,hamdi2024llm,ait2024contextual,antoun2020arabert}. Fine-tuning even on small labeled datasets improves performance and bidirectional context capture. Refinements like RoBERTa's NSP task removal boost domain-specific results, achieving an F1 score of 0.8 in medication detection and 3.929 MAE in eye-tracking tasks~\cite{b10,b12,b8}.

Efficient light-weight models like DistilBERT, which is 40\% smaller and 60\% faster, retain 97\% BERT's performance, excelling in tasks such as the classification of socio-political news \cite{b11,b8}. ALBERT-Base-v2 optimized memory and speed, achieving a 13.04 exact match score in COVID-19 queries~\cite{b8}. XLM-RoBERTa improved results in processing more than 100 languages using the advances in multilingual processing and outperforming prior models by 23\% in multilingual accuracy~\cite{b19,b14,b15,b8,b13}. Special adaptations like Electra-Small and BART-Large introduced token substitution and hybrid architectures. Electra-Small excelled in multilingual fake news detection with innovative token substitution strategies~\cite{b16,b8,abdellatif2024lmrpa}. Both models achieved high performance in specialized tasks such as medical complaint detection and NER, with BART-Large reducing voice recognition errors by 21.7\%~\cite{b18,b17,b8}. These advancements highlight LLMs' potential in handling noisy and unstructured text data.

Generative LLMs enhance NLP, excelling in specialized tasks. Fine-tuned models like GPT-3.5 and Mistral-7B surpass baselines by 50\% in F1-macro scores on domain-specific datasets~\cite{b7}. QLoRA improves memory efficiency~\cite{b6}, but model variability persists, as seen in GPT-NeoX-20B and Llama2-7B.

Preprocessing techniques such as text refinement, summarization, and Named Entity Recognition (NER), are essential for handling noisy and unstructured text~\cite{pennington2014glove}. BERT and RoBERTa have performed extremely well in NER tasks, effectively extracting key entities from complex text~\cite{b10,b8, devlin2019bert}. Combining preprocessing with fine-tuning enhances classification accuracy, especially when raw data lacks structure or clarity.

Based on this, our study introduces a multi-layered framework that uses LLM preprocessing to improve the fine-tuning of Arabic language models and address noise in data. This enhances the effectiveness of Arabic language models such as CAMeL-BERT, AraBERT, and Asafaya-BERT, in disease classification. It also presents an effective strategy for incorporating LLMs into social telehealth applications.

\section{Dataset}
The dataset used in this study was collected from user-generated posts on an online social platform where patients shared their medical complaints in Arabic. The posts contained detailed information such as the status of chronic diseases, descriptions of symptoms, symptom durations, height, weight, gender, and age, categorized into Type, and Diagnosis. Structuring and annotation were done under the supervision of a medical advisor to ensure relevance and accuracy.

\begin{figure*}[h]
    \centering
        
        \includegraphics[width=0.8\textwidth]{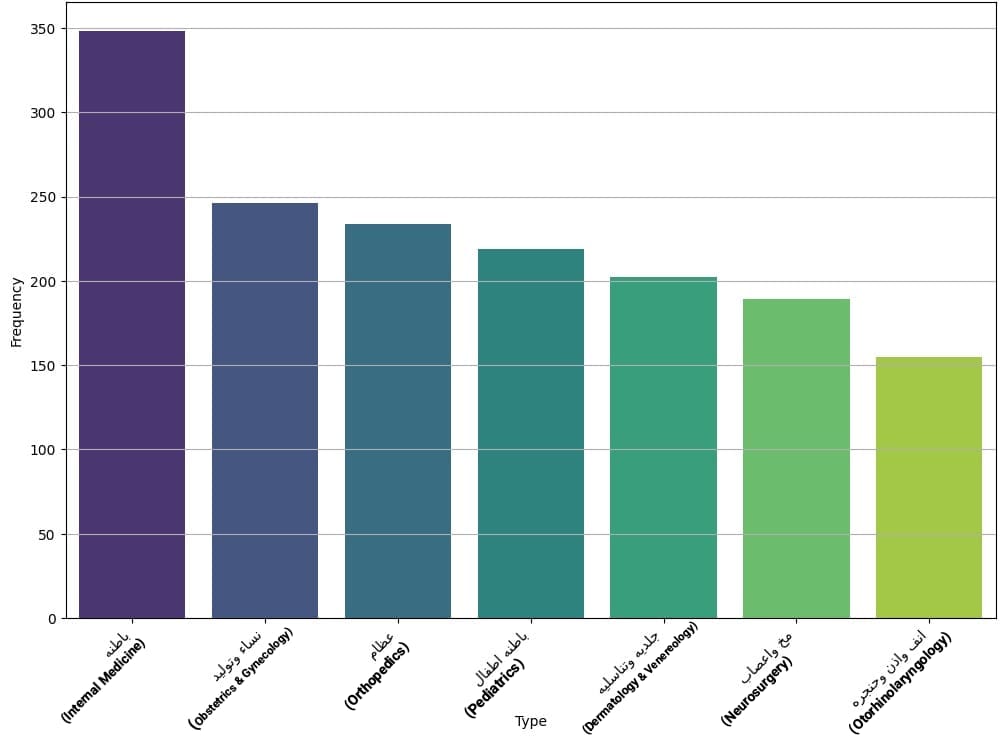}
        \caption{Distribution of disease types in the dataset.}
        \label{fig:condition_types}
 \end{figure*}

Figure \ref{fig:condition_types} shows the distribution of our dataset across seven disease types (classes) such as Internal Medicine, Orthopedics, Neurosurgery, etc., with Internal Medicine having the highest number of posts.

\section{Methodology}
This work proposes an Ensemble Classification on Arabic language model fine-tuning using a multi-layered framework. The LLAMA3 model has been used in the enhanced preprocessing step; the flow is illustrated as shown in Fig.~\ref{fig:image1}.

\begin{table*}[htbp]
\caption{Comparison of Text, Refined, Summarized, and NER}
\label{tab:comparison_full}
\centering
\renewcommand{\arraystretch}{1.4}
\begin{tabularx}{\textwidth}{|X|X|X|X|}
\hline
\textbf{Text} & \textbf{Refined} & \textbf{Summarized} & \textbf{NER} \\
\hline

\RL{السلام عليكم ورحمة الله وبركاته 
شاب 22 سنه بروح الجيم يوميا بس بحس أحيانا بضعف والم ف العضلات بعد التمرين 
محتاج دكتور مختص يرد عليا بعلاج مناسب لتقوية الأعصاب او فيتامينات لتقوية الأعصاب 
الوزن 55
الطول 169}
&
\RL{الشاب يعاني أحيانا بضعف عضلات بعد التمرين
الشاب يطلب العلاج من دكتور مختص لتقوية الأعصاب
الشاب يحتاج إلى فيتامينات لتقوية الأعصاب
الوزن 55
الطول 169}
&
\RL{ضعف العضلات بعد التمرين
العمر 22 سنة
الوزن 55 كجم
الطول 169 سم}
&
\RL{الضعف والم في العضلات بعد التمرين}
\\
\hline

\small{Peace be upon you. A 22-year-old young man goes to the gym daily but sometimes feels weakness and muscle pain after training. He needs a specialized doctor to advise on proper treatment to strengthen the nerves or vitamins for nerve strengthening. Weight 55, Height 169.}
&
\small{The young man sometimes suffers from muscle weakness after exercise. He seeks treatment from a specialist doctor to strengthen the nerves. He needs vitamins for nerve strengthening. Weight 55, Height 169.}
&
\small{Muscle weakness after exercise. Age 22 years. Weight 55 kg. Height 169 cm.}
&
\small{Weakness and muscle pain after exercise.}
\\
\hline

\end{tabularx}
\end{table*}

\subsection{Medical Questions}\label{AA}
The dataset contains Arabic text data from user-generated content on online health platforms. Texts include elaborate descriptions by the patients of their symptoms, patient medical history, age, and gender. While doing preprocessing, it automatically removes private and sensitive information to ensure privacy.

\vspace{0.3cm}

\subsection{Multi-layer preprocessing using LLAMA3}The text is further enhanced using a multi-layered approach to LLAMA 3 filtering. The process involved in preprocessing is as follows:

\begin{itemize}
    \item \textbf{Text Refinement:} LLAMA3 improves the text through deleting unmet requirements such as irrelevant information, grammatical mistakes and the vague parts of the material while protecting the medical context.
    \item \textbf{Text Summarization:}  Llama 3 eliminates unnecessary aspects of long medical posts and compresses them into summaries for easier understanding.
    \item \textbf{Named Entity Recognition (NER):} Symptoms and conditions, together with the drugs, are important medical entities and LLAMA 3 identifies and extracts them.
\end{itemize}

All these steps (refined text, summarized text, and NER-extracted entities) are used to augment the base data.

\subsection{Data Annotation}
To create new data variants, the performance data is combined with the outputs from the preprocessing stages. These enriched datasets are particularly effective for multi-label classification tasks. The raw text data generated by users is processed through three LLM-based preprocessing methods: refinement, summarization, and Named Entity Recognition (NER) as shown in Table~\ref{tab:comparison_full}.

To preserve contextual richness, the original text posts are separately combined with each preprocessed output, resulting in three paired representations:
\begin{itemize}
    \item (original + refined)
    \item (original + summarized)
    \item (original + NER)
\end{itemize}

Each paired dataset is then annotated and used to fine-tune Arabic language models.

\subsection{Arabic Language Models Fine-Tuning}
Three different pre-trained transformer models were fine-tuned individually for the classification task:
\begin{itemize}
    \item \textbf{CAMeL-Lab/bert-base-arabic-camelbert-mix}
    \item \textbf{aubmindlab/bert-base-arabert}
    \item \textbf{asafaya/bert-base-arabic}
\end{itemize}

Each model was adapted by adding a dropout layer followed by a linear classification head to output predictions over the seven categories. All models were trained under the same hyperparameter configuration to ensure a fair comparison. The training configuration was as follows:
\begin{itemize}
    \item Dropout Rate: 0.05
    \item Learning Rate: 1e-4
    \item Batch Size: 4
    \item Number of Epochs: 25
    \item Weight Decay: 0.01
    \item Loss Function: Cross-Entropy Loss
\end{itemize}

\subsection{Ensemble Learning: Majority Voting}
After fine-tuning the individual models, an ensemble strategy based on Majority Voting was applied. For each question in the test set, predictions from all three models were collected, and the final prediction was determined based on the majority class among the outputs. 

The ensemble approach aimed to combine the strengths of different models, reduce variance, and improve the overall prediction robustness and reliability.

\section{Results}
As illustrated in Table~\ref{tab:model_performance}, to evaluate the classification performance of different ensemble configurations, we constructed ensemble combinations from twelve base models:

\begin{itemize}
    \item \textbf{CAMeLBERT}: Post, Refined, NER, Summarized
    \item \textbf{AraBERT}: Post, Refined, NER, Summarized
    \item \textbf{AsafayaBERT}: Post, Refined, NER, Summarized
\end{itemize}

These models were selected due to their linguistic diversity and preprocessing variations, allowing us to test the effects of ensemble size and model variety.

\begin{table}[H]
\centering
\caption{Performance comparison between individual models and the ensemble strategy.}
\label{tab:model_performance}
\renewcommand{\arraystretch}{1.15}
\resizebox{0.82\linewidth}{!}{%
\begin{tabular}{lc}
\toprule
\textbf{Model} & \textbf{Accuracy} \\
\midrule
\multicolumn{2}{l}{\textbf{CAMeLBERT}} \\
CamelBERT\_Post & 70.53\% \\
CamelBERT\_Refined & 75.55\%  \\
CamelBERT\_NER & 74.61\%  \\
CamelBERT\_Summarized & 64.89\%  \\
\midrule
\multicolumn{2}{l}{\textbf{AraBERT}} \\
AraBERT\_Post & 71.79\%  \\
AraBERT\_Refined & 72.41\%  \\
AraBERT\_NER & 68.97\%  \\
AraBERT\_Summarized & 65.94\%  \\
\midrule
\multicolumn{2}{l}{\textbf{AsafayaBERT}} \\
AsafayaBERT\_Post & 74.92\%  \\
AsafayaBERT\_Refined & 75.24\%  \\
AsafayaBERT\_NER & 73.67\%  \\
AsafayaBERT\_Summarized & 74.92\%  \\
\midrule
\textbf{Ensemble (Majority Voting)} & \textbf{80.56\%} \\
\bottomrule
\end{tabular}%
}
\end{table}

\begin{figure*}[htbp]
\centering
\begin{minipage}[t]{0.48\textwidth}
    \includegraphics[width=\linewidth]{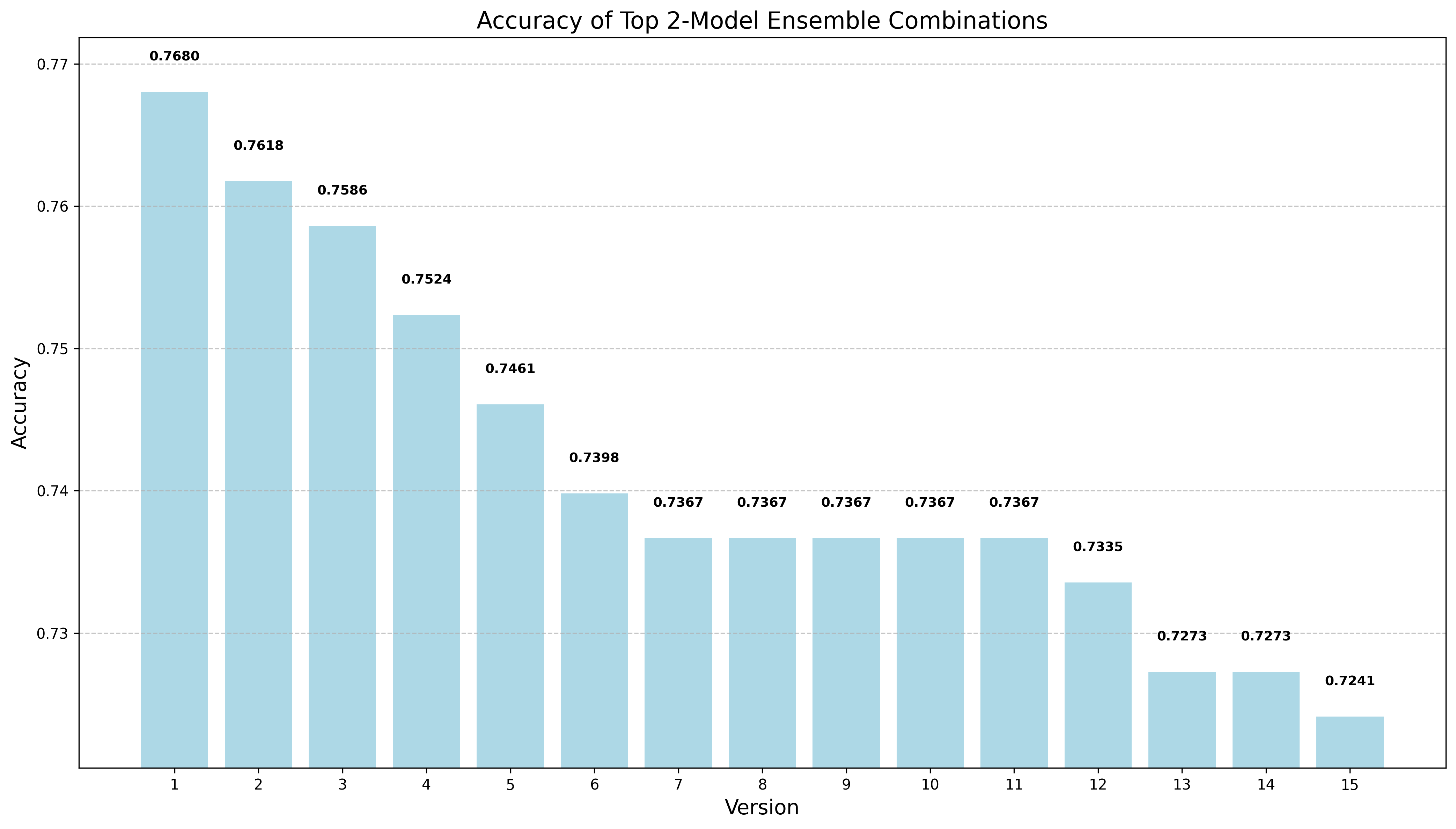}
    \caption{Accuracy of Top 2-Model Ensemble Combinations. The best combination achieved an accuracy of 0.7680, showing limited performance improvement with only two models.}
    \label{fig:top2ensemble}
\end{minipage}
\hfill
\begin{minipage}[t]{0.48\textwidth}
    \includegraphics[width=\linewidth]{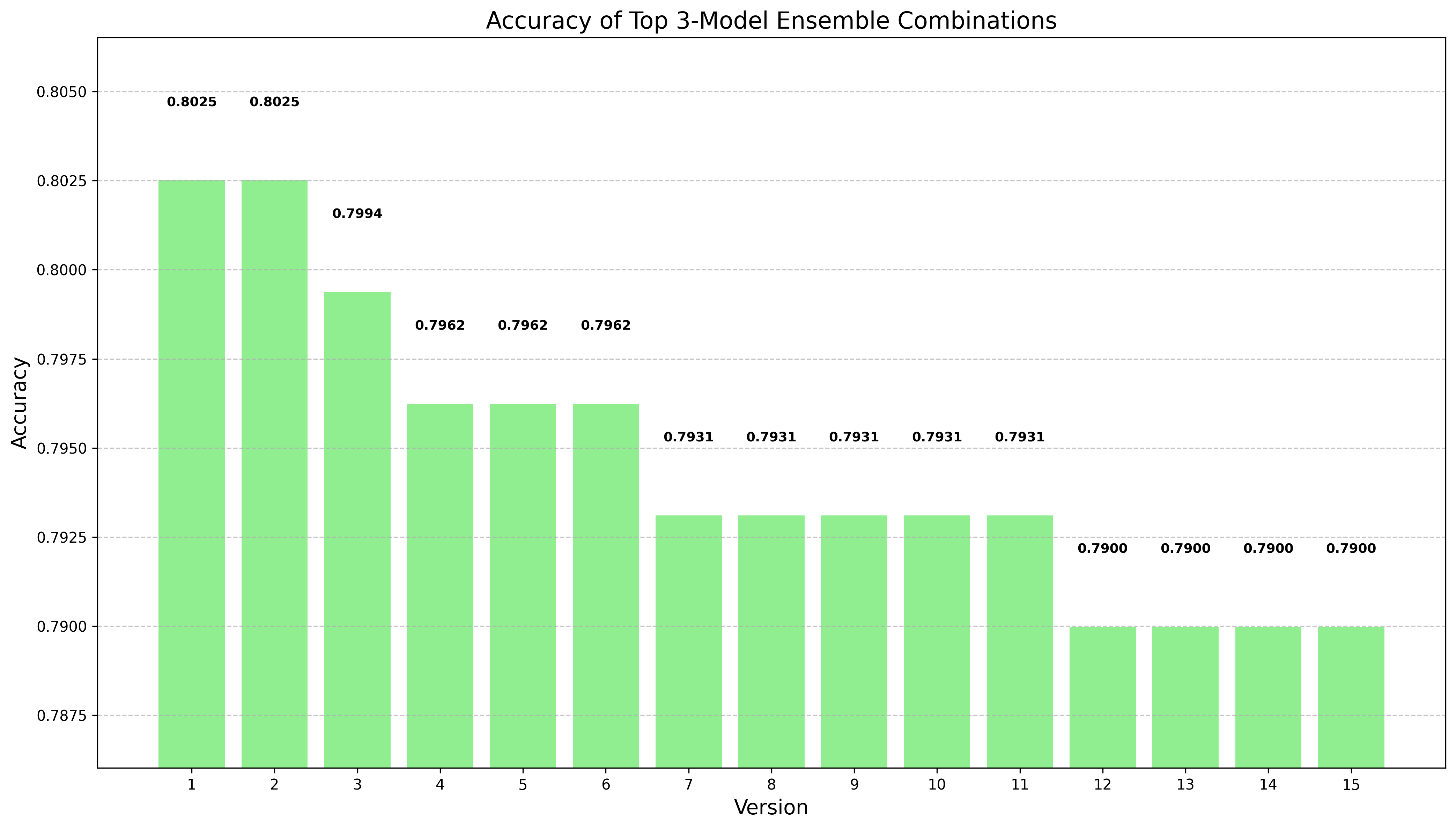}
    \caption{Accuracy of Top 3-Model Ensemble Combinations. Accuracy improves significantly to 0.8025, indicating the benefits of model diversity.}
    \label{fig:top3ensemble}
\end{minipage}
\end{figure*}

\begin{figure*}[htbp]
\centering
\begin{minipage}[t]{0.48\textwidth}
    \includegraphics[width=\linewidth]{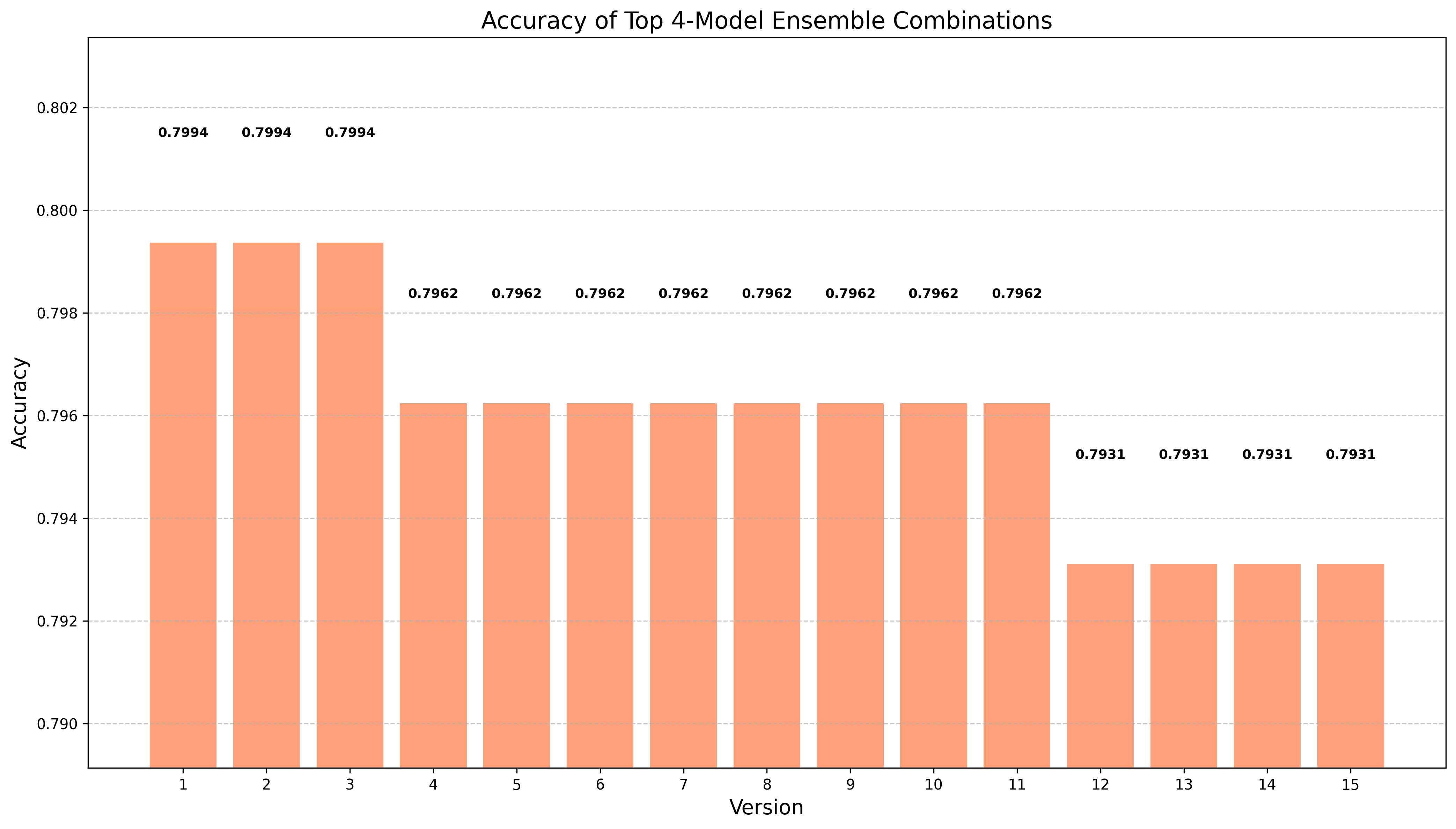}
    \caption{Accuracy of Top 4-Model Ensemble Combinations. The best ensemble achieved 0.7994, with diminishing returns beyond three models.}
    \label{fig:top4ensemble}
\end{minipage}
\hfill
\begin{minipage}[t]{0.48\textwidth}
    \includegraphics[width=\linewidth]{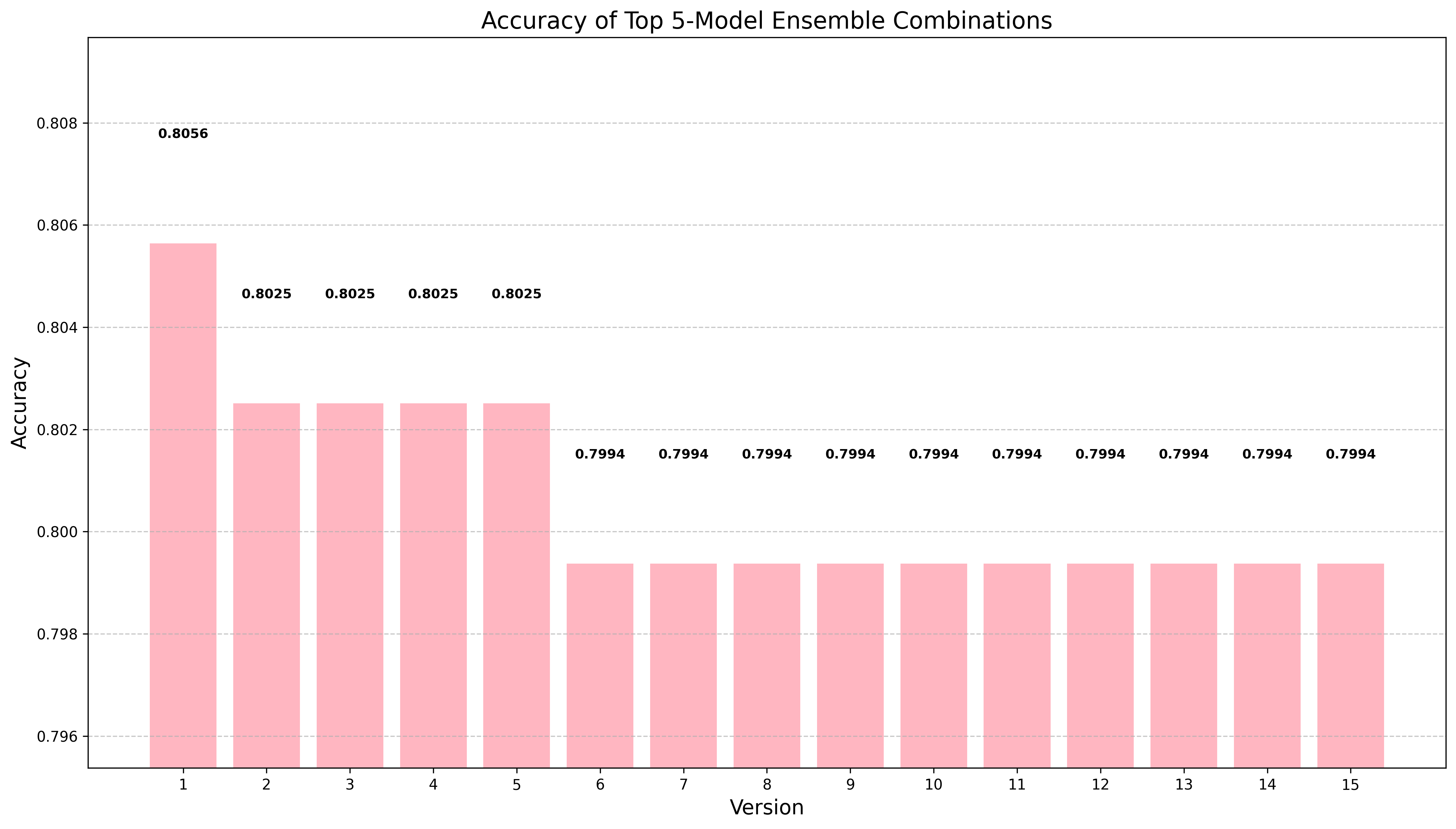}
    \caption{Accuracy of Top 5-Model Ensemble Combinations. Reaches a peak of 0.8056, one of the best overall performances across all configurations.}
    \label{fig:top5ensemble}
\end{minipage}
\end{figure*}

\begin{figure*}[htbp]
\centering
\begin{minipage}[t]{0.48\textwidth}
    \includegraphics[width=\linewidth]{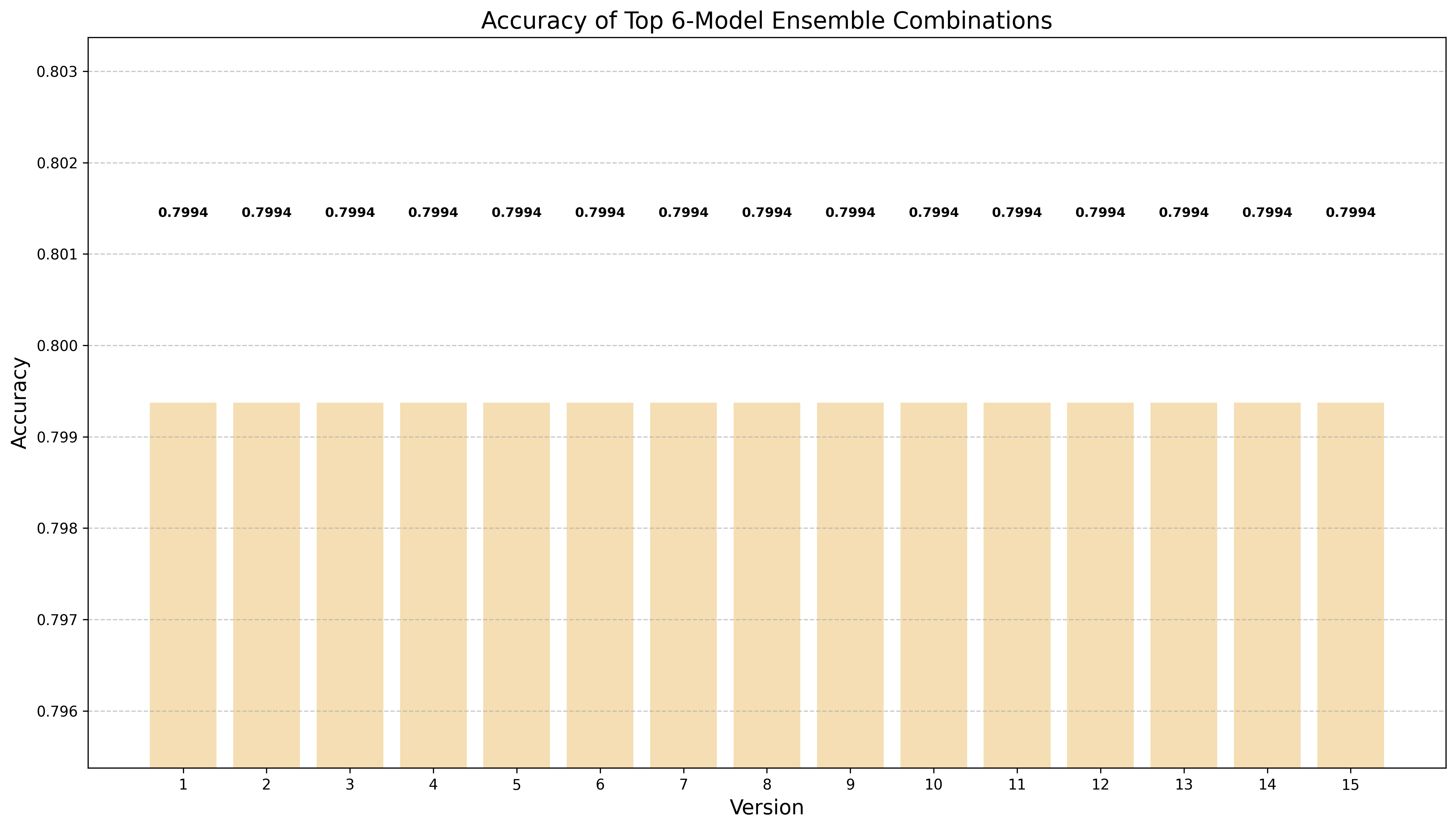}
    \caption{Accuracy of Top 6-Model Ensemble Combinations. All combinations plateaued at 0.7994, suggesting ensemble saturation.}
    \label{fig:top6ensemble}
\end{minipage}
\hfill
\begin{minipage}[t]{0.48\textwidth}
    \includegraphics[width=\linewidth]{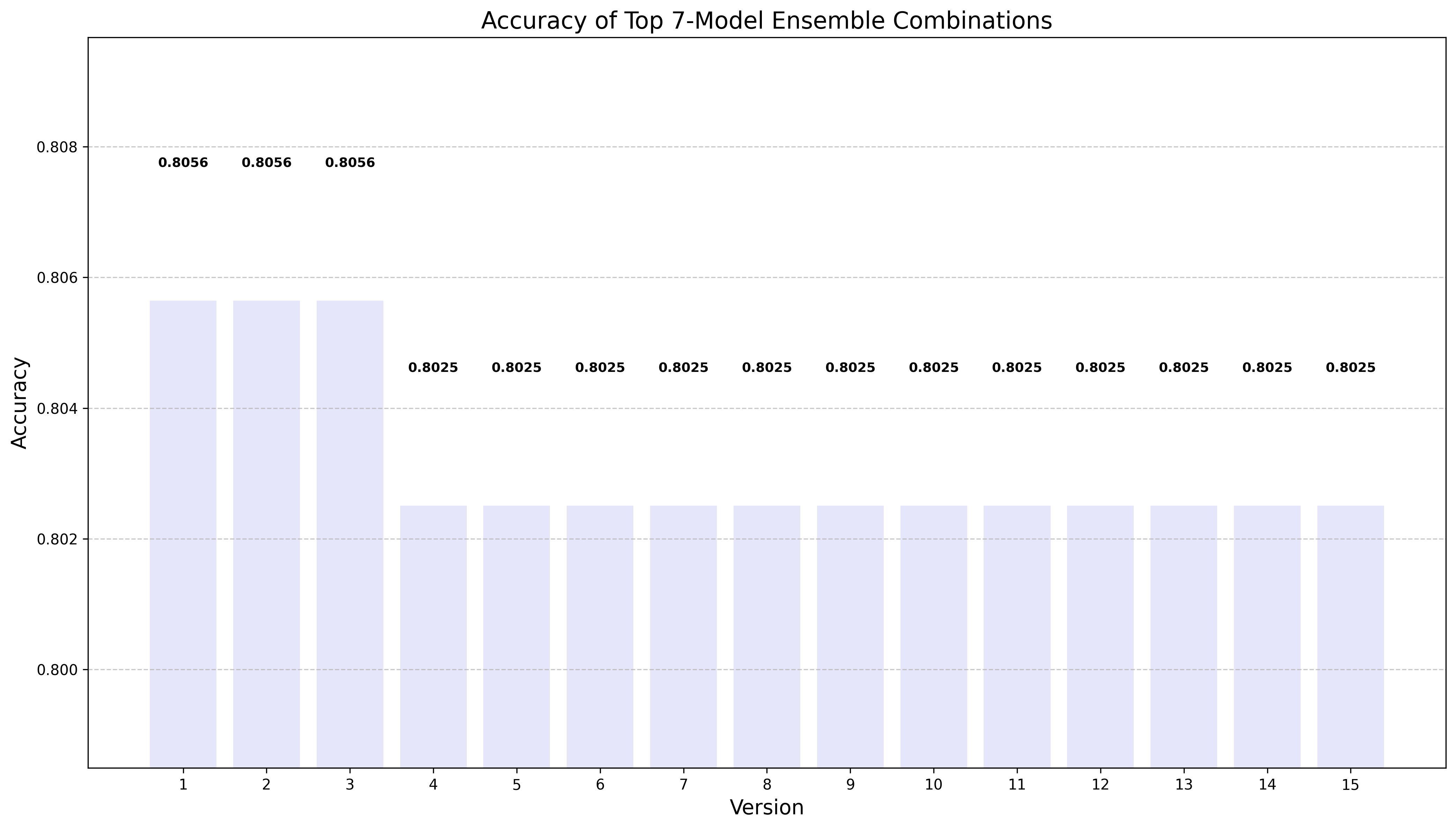}
    \caption{Accuracy of Top 7-Model Ensemble Combinations. The highest accuracy of 0.8056 indicates effective scaling with selected models.}
    \label{fig:top7ensemble}
\end{minipage}
\end{figure*}

\begin{figure*}[htbp]
\centering
\begin{minipage}[t]{0.48\textwidth}
    \includegraphics[width=\linewidth]{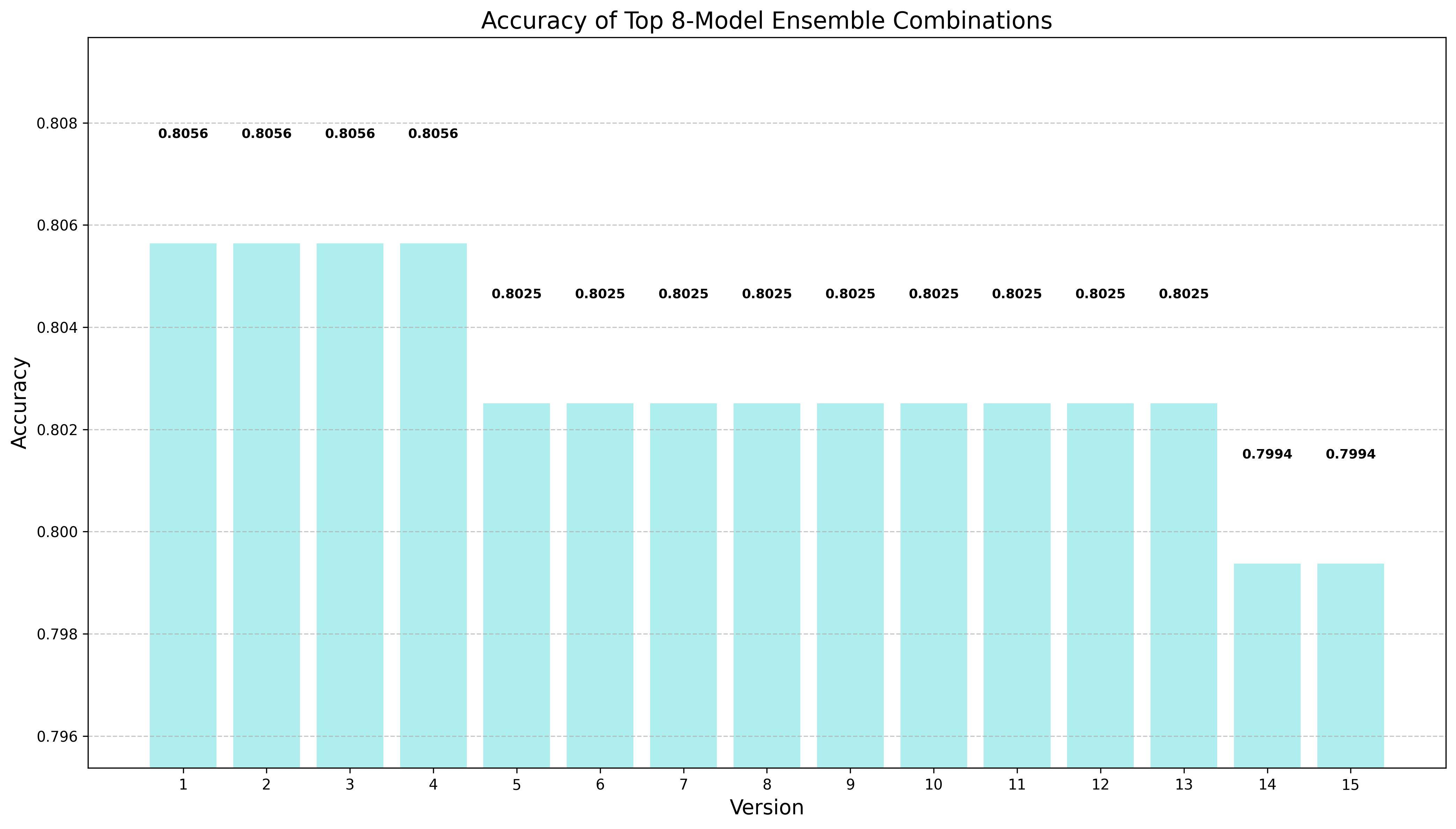}
    \caption{Accuracy of Top 8-Model Ensemble Combinations. Matches best results at 0.8056, maintaining high performance.}
    \label{fig:top8ensemble}
\end{minipage}
\hfill
\begin{minipage}[t]{0.48\textwidth}
    \includegraphics[width=\linewidth]{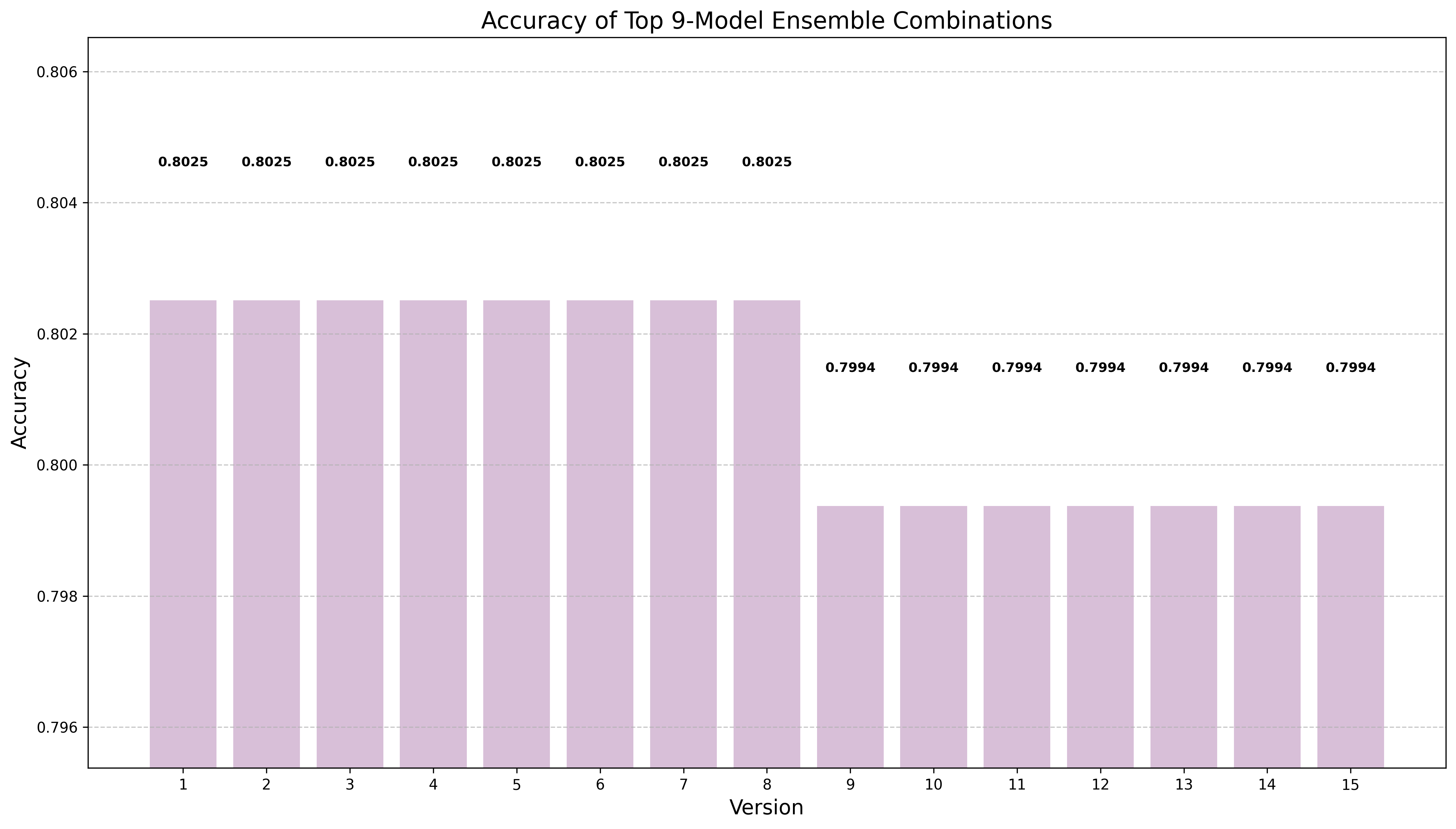}
    \caption{Accuracy of Top 9-Model Ensemble Combinations. Peak accuracy is slightly lower at 0.8025, indicating diminishing benefits.}
    \label{fig:top9ensemble}
\end{minipage}
\end{figure*}

\begin{figure*}[htbp]
\centering
\begin{minipage}[t]{0.48\textwidth}
    \includegraphics[width=\linewidth]{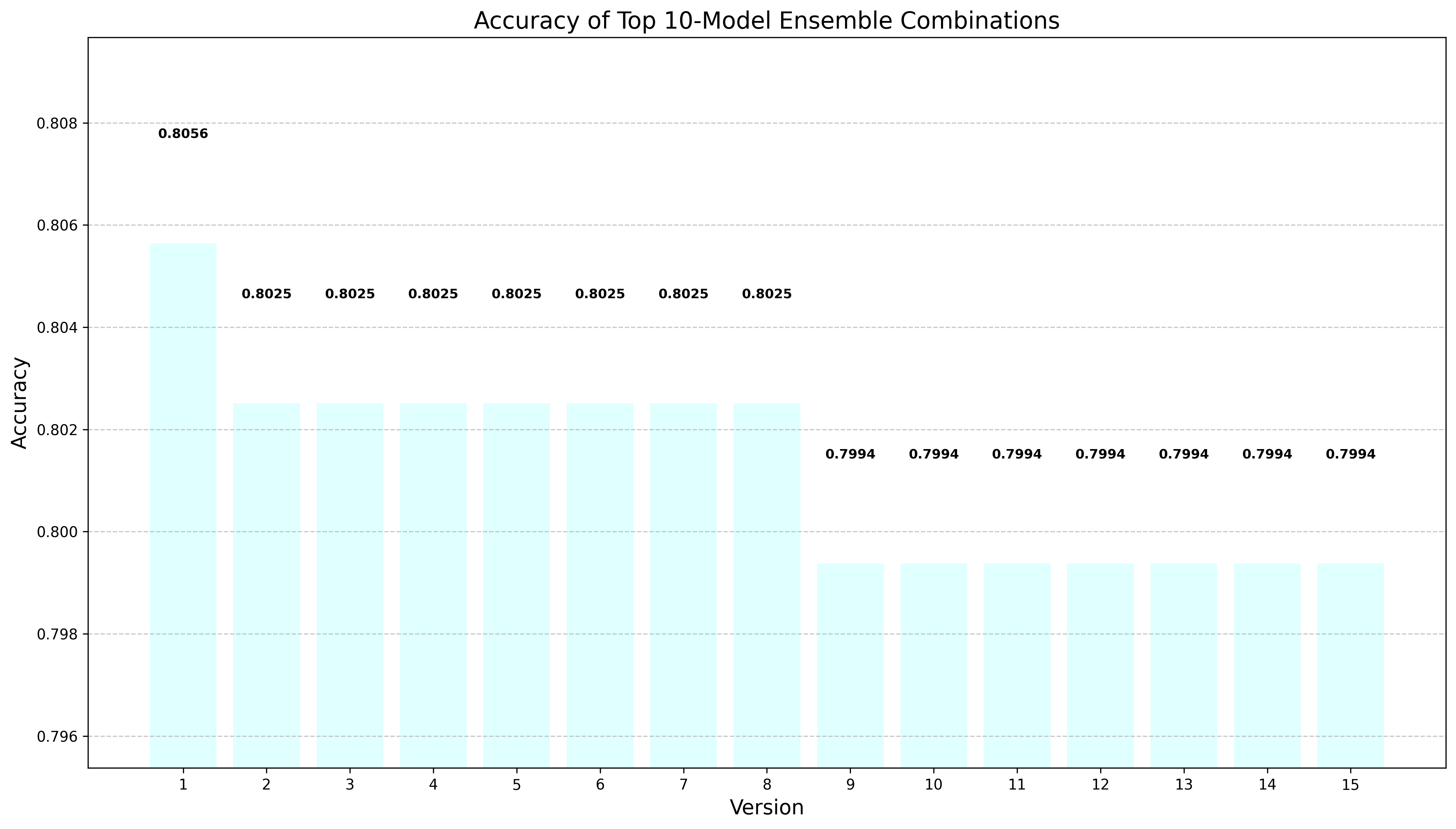}
    \caption{Accuracy of Top 10-Model Ensemble Combinations. Matches the highest accuracy of 0.8056, showing that larger ensembles can still be optimal.}
    \label{fig:top10ensemble}
\end{minipage}
\hfill
\begin{minipage}[t]{0.48\textwidth}
    \includegraphics[width=\linewidth]{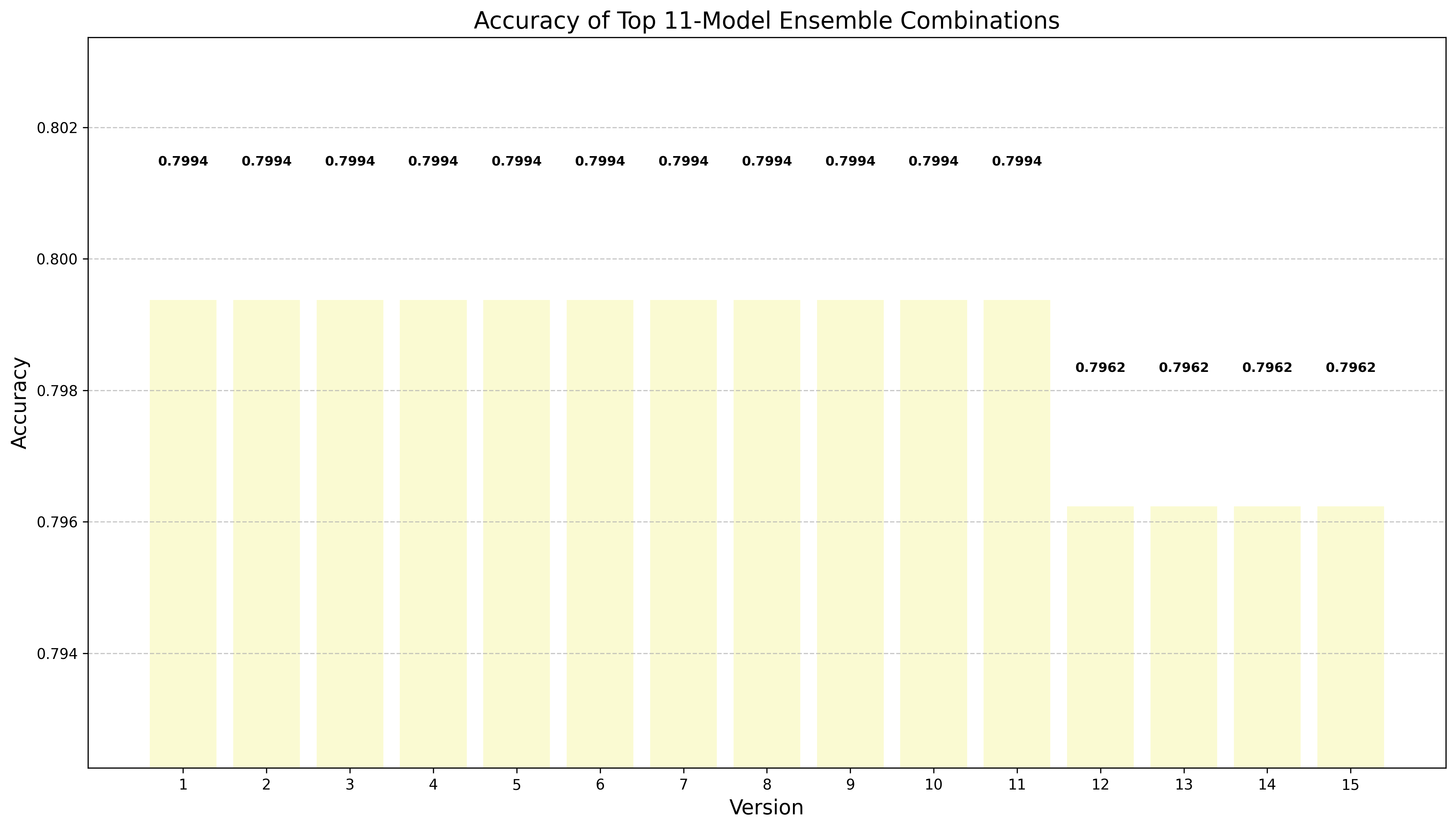}
    \caption{Accuracy of Top 11-Model Ensemble Combinations. Maximum accuracy observed was 0.7994, with reduced variation across versions.}
    \label{fig:top11ensemble}
\end{minipage}
\end{figure*}

\begin{figure*}[htbp]
\centering
\begin{minipage}[t]{0.48\textwidth}
    \centering
    \includegraphics[width=\linewidth,height=5.5cm]{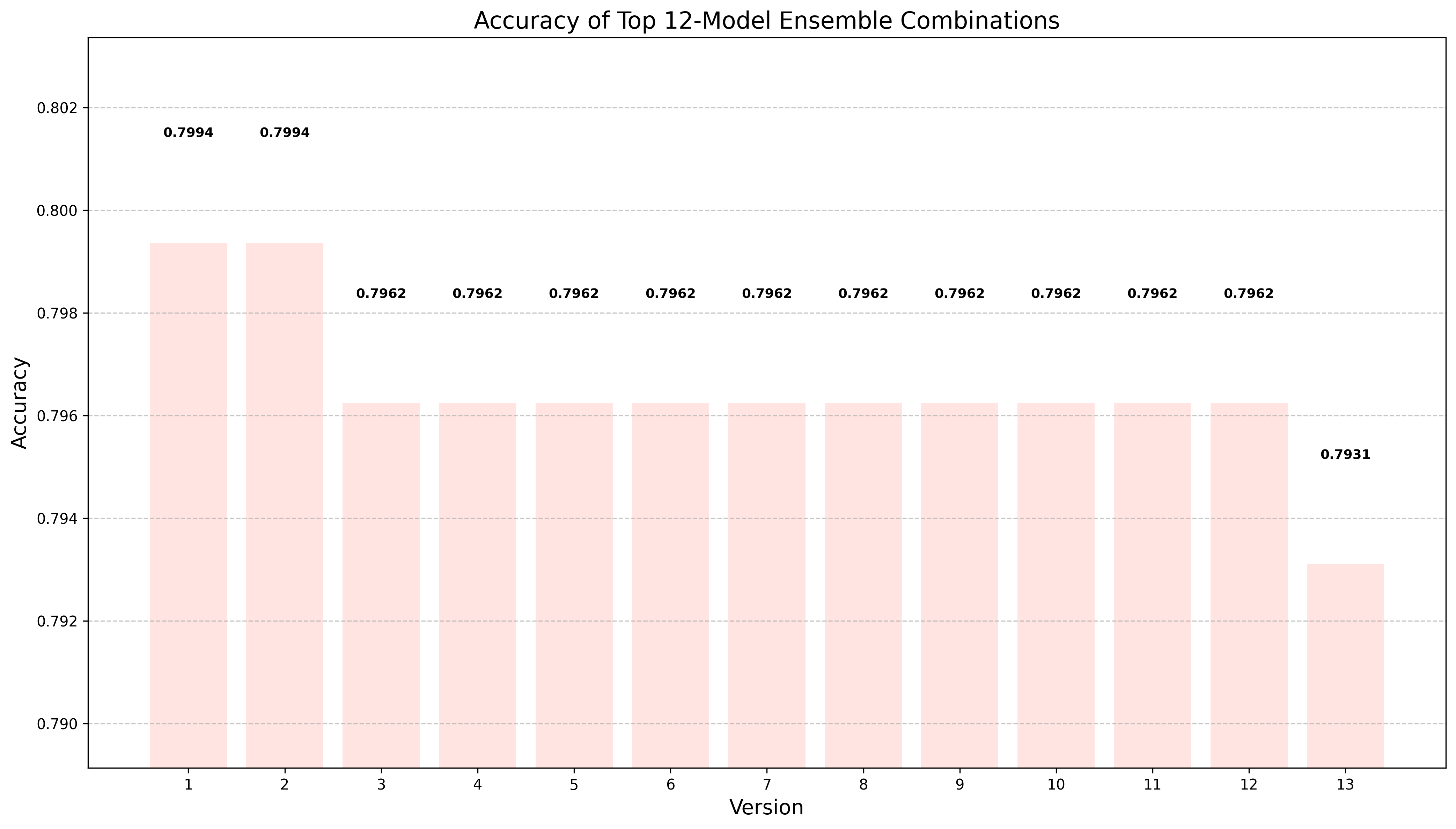}
    \caption{Accuracy of Top 12-Model Ensemble Combinations. The highest recorded was 0.7994.}
    \label{fig:top12ensemble}
\end{minipage}
\hfill
\begin{minipage}[t]{0.48\textwidth}
    \centering
    \includegraphics[width=\linewidth,height=5.5cm]{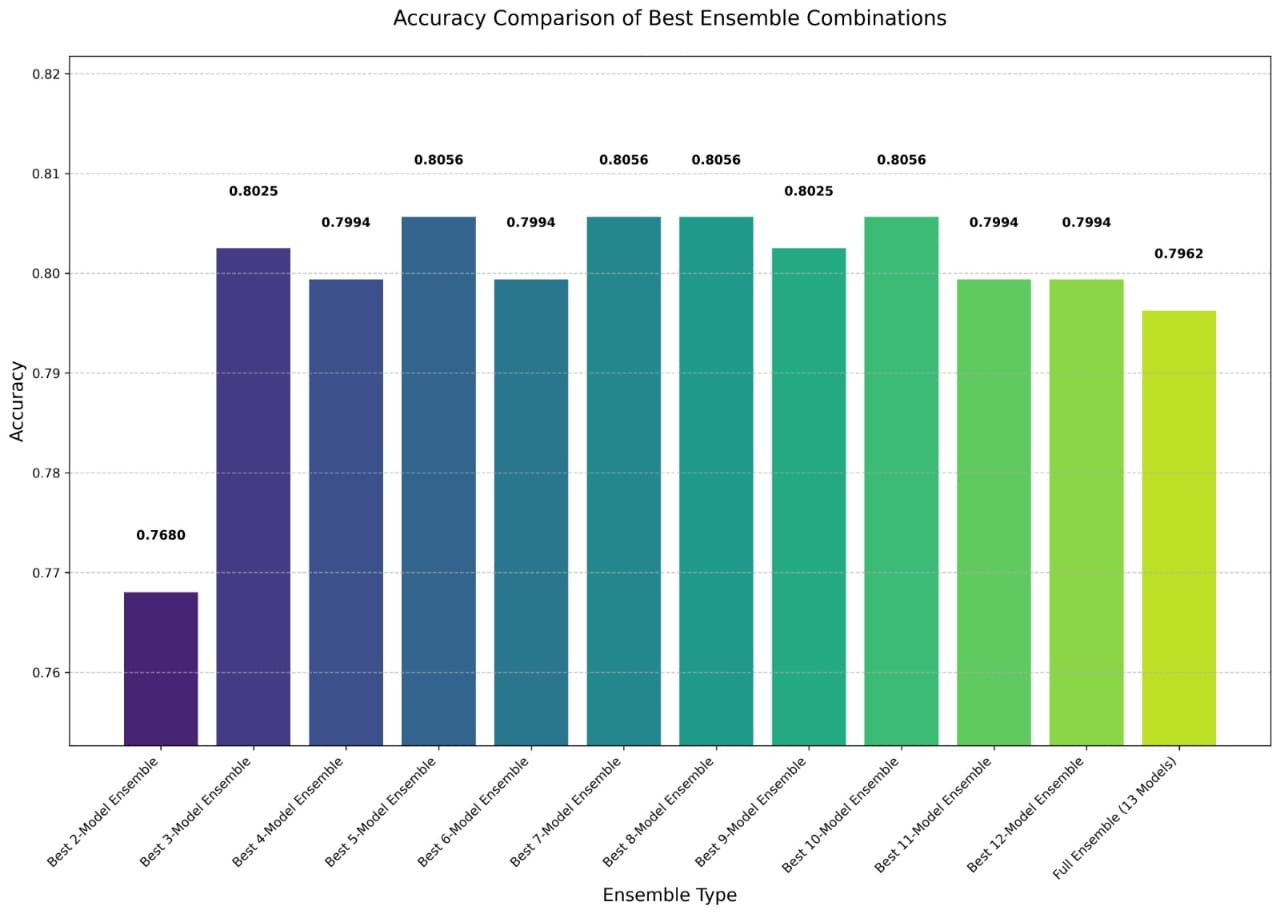}
    \caption{Comparison of Best Ensemble Combinations Across Different Ensemble Sizes. The highest recorded accuracy was 0.8056, observed consistently across the 4, 6, 7, and 10-model ensembles.}
    \label{fig:ensemblecomparison}
\end{minipage}
\end{figure*}

\begin{figure*}[htbp]
\centering
\includegraphics[width=0.8\textwidth]{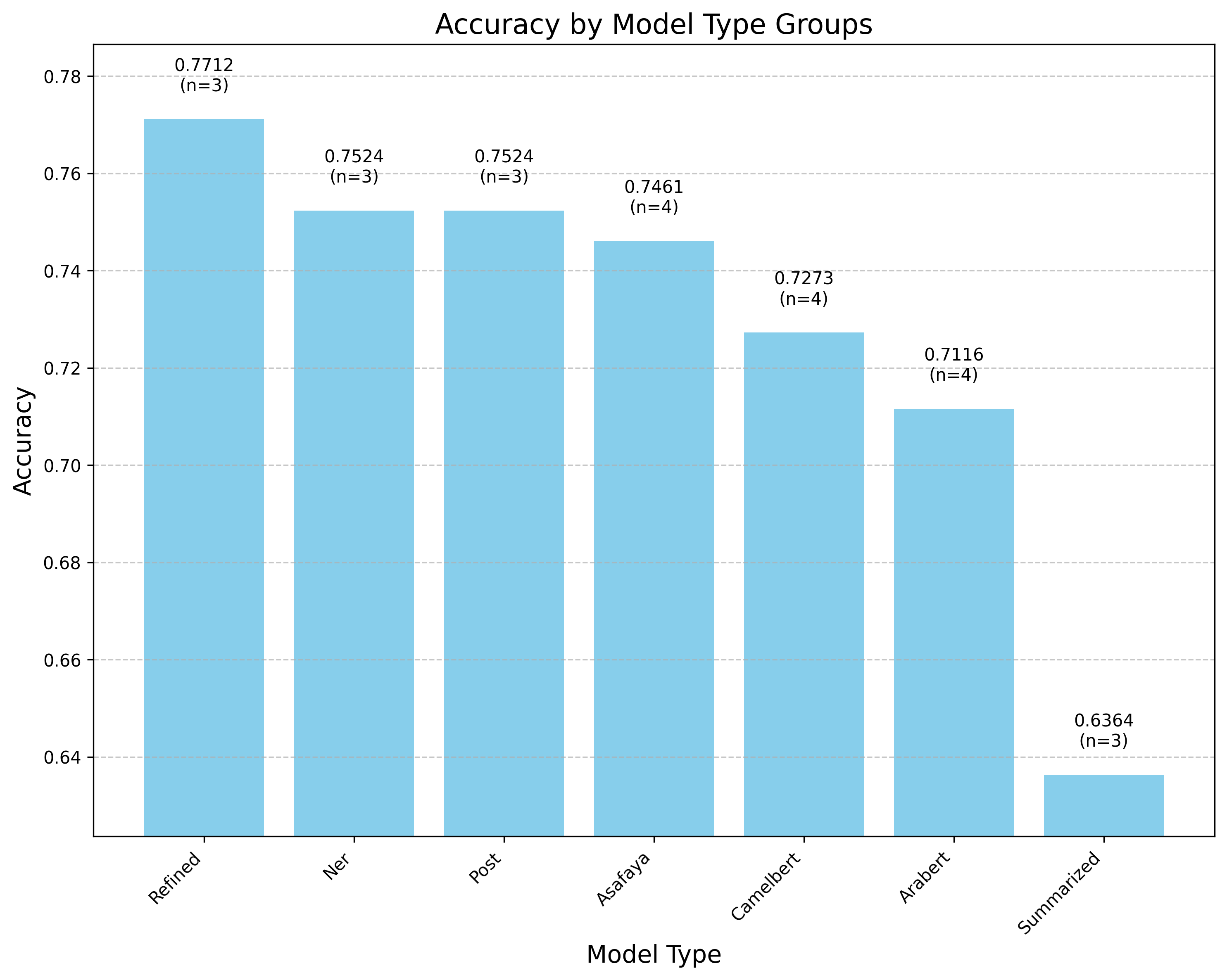}
\caption{Accuracy by Model Type Groups. The ensemble with refined models achieved the highest accuracy, while summarized models performed the worst.}
\label{fig:grouped_accuracy}
\end{figure*}

\noindent
Overall, the results reveal that ensemble strategies improve disease classification performance when multiple diverse models are combined. However, the accuracy gains diminish beyond three models unless the ensemble is carefully curated. The best accuracy of 0.8056 was consistently observed across ensemble sizes of 5, 7, 8, and 10 models.

To further analyze ensemble performance, we conducted experiments by grouping models based on two criteria: first, the preprocessing augmentation type (Refined, NER, Post, Summarized), and second, the underlying transformer architecture (AraBERT, CAMeLBERT, AsafayaBERT). For each group, we created an ensemble by selecting all models belonging to that category and applied a majority voting strategy to generate final predictions.

As illustrated in Figure~\ref{fig:grouped_accuracy}, the ensemble constructed using Refined models achieved the highest accuracy at 0.7712, followed closely by the NER and Post-based ensembles, both reaching 0.7524. In contrast, the ensemble using Summarized models performed the worst with an accuracy of 0.6364, indicating that excessive compression of medical content may hinder classification performance.

In terms of base models, the ensemble composed of AsafayaBERT variants outperformed those formed from CAMeLBERT and AraBERT models. The AraBERT-based ensemble recorded the lowest accuracy (0.7116) among the model-type groups. These results suggest that both the choice of preprocessing strategy and the underlying language model architecture play a critical role in ensemble-based classification accuracy.

This experiment highlights the benefit of majority voting ensembles that are strategically grouped, either by augmentation method or model family, for improving disease classification from Arabic medical texts.

\section{Conclusion}
This work contributes in combining fine-tuned the Arabic language models for the purpose of disease classification in social telehealth applications. The methodology involving multi-layered preprocessing for text Refinement, Summarization, and Named Entity Recognition (NER) to extract key words of medical posts. The enriched datasets were used to fine-tune three notable Arabic language models: CAMeL-BERT, AraBERT, and Asafaya-BERT, combining them with majority voting ensembling technique to enhance the accuracya and the best performance achieved 80.56\%. This approach improves performance and provides a new benchmark for its application in telehealth. Future studies can extend this framework to other languages or combining
multiple models using an ensemble technique for severity assessment.


\begin{thebibliography}{00}

\bibitem{b1} J. Chen and Y. Wang, ``Social media use for health purposes: Systematic review,'' \textit{Journal of Medical Internet Research}, vol. 23, no. 5, 2020.

\bibitem{b2} R. B. Correia, I. B. Wood, J. Bollen, and L. M. Rocha, ``Mining social media data for biomedical signals and health-related behavior,'' \textit{Annual Review of Biomedical Data Science}, vol. 3, no. 1, pp. 433--458, 2020.

\bibitem{b3} A. Magge \textit{et al.}, ``Overview of the Sixth Social Media Mining for Health Applications (SMM4H) Shared Tasks at NAACL 2021,'' in \textit{Proc. of the Sixth Social Media Mining for Health Applications (SMM4H) Shared Task at NAACL}, Jan. 2021.

\bibitem{b4} Y. Guo, A. Ovadje, M. A. Al-Garadi, and A. Sarker, ``Evaluating large language models for health-related text classification tasks with public social media data,'' \textit{Journal of the American Medical Informatics Association}, 2024.

\bibitem{b5} K. Kowsari, K. J. Meimandi, M. Heidarysafa, S. Mendu, L. Barnes, and D. Brown, ``Text classification algorithms: A survey,'' \textit{Information}, vol. 10, no. 4, p. 150, Apr. 2019.

\bibitem{b6} S. S. Alahmari, L. O. Hall, P. R. Mouton, and D. B. Goldgof, ``Repeatability of fine-tuning large language models illustrated using QLORA,'' \textit{IEEE Access}, 2024.

\bibitem{b7} N. Niraula, S. Ayhan, B. Chidambaram, and D. Whyatt, ``Multi-label classification with generative large language models,'' in \textit{Proc. of the 43rd Digital Avionics Systems Conference (DASC)}, 2024.

\bibitem{b8} R. Qasim, W. H. Bangyal, M. A. Alqarni, and A. A. Almazroi, ``A fine-tuned BERT-based transfer learning approach for text classification,'' \textit{Journal of Healthcare Engineering}, 2022.

\bibitem{b9} V. Radivchev and A. Nikolov, ``Nikolov-Radivchev at SemEval-2019 Task 6: Offensive tweet classification with BERT and ensembles,'' in \textit{Proc. of the 13th Int. Workshop on Semantic Evaluation (SemEval-2019)}, pp. 691--695, 2019.

\bibitem{b10} S. Casola and A. Lavelli, ``FBK @ SMM4H 2020: RoBERTa for detecting medications on Twitter,'' in \textit{Proc. of the Fifth Social Media Mining for Health Applications Workshop \& Shared Task}, pp. 101--103, 2020.

\bibitem{b11} B. Büyüköz, A. Hürriyetoğlu, and A. Özgür, ``Analyzing ELMo and DistilBERT on socio-political news classification,'' in \textit{Proc. of the Workshop on Automatic Extraction of Socio-political Events from News}, pp. 9--18, USA, May 2020.

\bibitem{b12} B. Li and F. Rudzicz, ``TorontoCL at CMCL 2021 shared task: RoBERTa with multi-stage fine-tuning for eye-tracking prediction,'' in \textit{Proc. of the Workshop on Cognitive Modeling and Computational Linguistics (CMCL)}, pp. 4--9, 2021.

\bibitem{b13} Y. Zhao and X. Tao, ``ZYJ123@DravidianLangTech-EACL2021: Offensive language identification based on XLM-RoBERTa with DPCNN,'' in \textit{Proc. of the First Workshop on Speech and Language Technologies for Dravidian Languages}, pp. 216--221, Parkville, Victoria: EACL, 2021.

\bibitem{b14} X. Ou and H. Li, ``YNU @ Dravidian-CodeMix-FIRE2020: XLM-RoBERTa for multi-language sentiment analysis,'' in \textit{Proc. of the Forum for Information Retrieval Evaluation (FIRE)}, pp. 4--9, 2020.

\bibitem{b15} O. Hamad, K. Shaban, and A. Hamdi, ``ASEM: Enhancing empathy in chatbot through attention-based sentiment and emotion modeling,'' in \textit{Proc. of the Int. Conf. on Computational Linguistics (COLING)}, pp. 1588--1601, 2024.

\bibitem{b16} K. A. Das, A. Baruah, F. A. Barbhuiya, and K. Dey, ``Ensemble of ELECTRA for profiling fake news spreaders,'' in \textit{Proc. of the Forum for Information Retrieval Evaluation (FIRE)}, pp. 22--25, 2020.

\bibitem{b17} C. Popa and T. Rebedea, ``BART-TL: Weakly-supervised topic label generation,'' in \textit{Proc. of the 16th Conf. of the European Chapter of the ACL (EACL)}, pp. 1418--1425, 2021.

\bibitem{b18} A. Mustar, S. Lamprier, and B. Piwowarski, ``Using BERT and BART for query suggestion,'' in \textit{Proc. of the CEUR Workshop on NLP}, vol. 2621, 2020.

\bibitem{b19} M. de Bruyn, E. Lotfi, J. Buhmann, and W. Daelemans, ``BART for knowledge grounded conversations,'' in \textit{Proc. of the CEUR Workshop on NLP}, vol. 2666, 2020.

\bibitem{abdellatif2024lmrpa} O. H. Abdellatif, A. N. Hassan, and A. Hamdi, ``LMRPA: Large language model-driven efficient robotic process automation for OCR,'' in \textit{The Int. Conf. of Advanced Computing and Informatics}, pp. 35--44, Springer, 2024.

\bibitem{hamdi2024llm} A. Hamdi, A. A. Mazrou, and M. Shaltout, ``LLM-SEM: A sentiment-based student engagement metric using LLMs for e-learning platforms,'' in \textit{The Int. Conf. of Advanced Computing and Informatics}, pp. 145--154, Springer, 2024.

\bibitem{mohamed2025llm} M. Mohamed, R. Emad, and A. Hamdi, ``A multi-layered large language model framework for disease prediction,'' \textit{arXiv preprint arXiv:2502.00063}, 2025.

\bibitem{devlin2019bert} J. Devlin, M.-W. Chang, K. Lee, and K. Toutanova, ``BERT: Pre-training of deep bidirectional transformers for language understanding,'' in \textit{Proc. of the 2019 Conf. of the NAACL}, pp. 4171--4186, 2019.

\bibitem{ait2024contextual} I. Ait Talghalit, H. Alami, and S. Ouatik El Alaoui, ``Contextual semantic embeddings based on transformer models for Arabic biomedical questions classification,'' \textit{HighTech and Innovation Journal}, vol. 5, no. 4, pp. 1024, 2024.

\bibitem{hamad2022steducov} O. Hamad, A. Hamdi, S. Hamdi, and K. Shaban, ``StEduCov: An explored and benchmarked dataset on stance detection in tweets towards online education during COVID-19 pandemic,'' \textit{Big Data and Cognitive Computing}, vol. 6, no. 3, pp. 88, 2022.

\bibitem{alali2025claseg} A. Al-Ali, A. Hamdi, M. Elshrif, \textit{et al.}, ``CLASEG: Advanced multiclassification and segmentation for differential diagnosis of oral lesions using deep learning,'' \textit{Scientific Reports}, vol. 15, no. 1, pp. 23016, 2025.

\bibitem{11078357} A. Allam, S. Ahmed, A. Hamdi, and A. Mohammed, ``Arabic large language models for medical text generation,'' in \textit{2025 4th Int. Conf. on Computer Technologies (ICCTech)}, pp. 1--6, 2025.

\bibitem{pennington2014glove} J. Pennington, R. Socher, and C. D. Manning, ``GloVe: Global vectors for word representation,'' in \textit{Proc. of the 2014 Conf. on Empirical Methods in NLP (EMNLP)}, pp. 1532--1543, 2014.

\bibitem{antoun2020arabert} W. Antoun, F. Baly, and H. Hajj, ``AraBERT: Transformer-based model for Arabic language understanding,'' in \textit{Proc. of the LREC 2020 Workshop on Resources for African Indigenous Languages}, pp. 9--15, 2020.

\bibitem{wei2019eda} J. Wei and K. Zou, ``EDA: Easy data augmentation techniques for boosting performance on text classification tasks,'' \textit{Proc. of the 2019 Conf. on Empirical Methods in NLP (EMNLP)}, pp. 6382--6388, 2019.

\end{thebibliography}
\end{document}